%% file: main.tex
\documentclass[journal]{IEEEtran}
\usepackage{xcolor}

\usepackage{multirow}
\usepackage{multicol}
\usepackage{booktabs}
\usepackage{soul}
\usepackage{hyperref}

\usepackage{cite}

\ifCLASSINFOpdf
   \usepackage[pdftex]{graphicx}
\else
\fi
\usepackage{amsmath}

\usepackage{stfloats}
\usepackage{url}

\begin{document}
%
\title{DZip: improved neural network based general-purpose lossless compression}
%
%
%

\author{Mohit~Goyal,
        Kedar~Tatwawadi, 
        Shubham~Chandak, 
        Idoia~Ochoa 
\thanks{M. Goyal and I. Ochoa are with the Department
of Electrical and Computer Engineering, University of Illinois at Urbana Champaign, 
IL, 61801 USA, correspondence e-mail: (mohit@illinois.edu). I. Ochoa is also with the School of Engineering at TECNUN, University of Navarra, Spain.}
\thanks{K. Tatwawadi and S. Chandak are with Department of Electrical Engineering, Stanford University, CA, USA.}
\thanks{Manuscript received April 19, 2005; revised August 26, 2015.}}

%
%

\markboth{Journal of \LaTeX\ Class Files,~Vol.~14, No.~8, August~2015}%
{Shell \MakeLowercase{\textit{et al.}}: Bare Demo of IEEEtran.cls for IEEE Journals}
%



\maketitle

\begin{abstract} 
We consider lossless compression based on statistical data modeling followed by prediction-based encoding, where an accurate statistical model for the input data leads to substantial improvements in compression. We propose DZip, a general-purpose compressor for sequential data that exploits the well-known modeling capabilities of neural networks (NNs) for prediction, followed by arithmetic coding. DZip uses a novel hybrid architecture based on adaptive and semi-adaptive training. Unlike most NN based compressors, DZip does not require additional training data and is not restricted to specific data types. The proposed compressor outperforms general-purpose compressors such as Gzip (29\% size reduction on average) and 7zip (12\% size reduction on average) on a variety of real datasets, achieves near-optimal compression on synthetic datasets, and performs close to specialized compressors for large sequence lengths, without any human input. While the main limitation of DZip in its current implementation is the encoding/decoding speed, we empirically demonstrate that DZip achieves comparable compression ratio to other NN-based compressors, while being several times faster. These results showcase the potential of developing improved general-purpose compressors based on neural networks and hybrid modeling. The source code for DZip and links to the datasets are available at \url{https://github.com/mohit1997/Dzip-torch}.
\end{abstract}


%
\IEEEpeerreviewmaketitle

\section{Introduction}
There has been a tremendous surge in the amount of data generated in the past years. Along with image and textual data, new types of data such as genomic, 3D VR, and point cloud data are being generated at a rapid pace \cite{1kgenome,pointcloud}. Thus, data compression is critical for reducing the storage and transmission costs associated with these data, and has been studied extensively from both theoretical and practical standpoints. In particular, a wide class of (lossless) compressors utilize the ``prediction + entropy coding'' approach, wherein a statistical model generates predictions for the upcoming symbols given the past and an entropy coder (e.g., arithmetic coder \cite{Witten:1987:ACD:214762.214771}) uses the predicted probabilities to perform compression. In this general framework, a better prediction model directly induces a better compressor. 

Given the close link between prediction and compression, there has been growing interest in using neural networks (NN) for compression due to their exceptional performance on several modeling and prediction tasks (e.g., language modeling \cite{devlin2018bert,radford2019language_gpt2} and generative modeling \cite{salimans2017pixelcnn++}). Neural network based models can typically learn highly complex patterns in the data much better than traditional finite context and Markov models, leading to significantly lower prediction error (measured as log-loss or perplexity \cite{devlin2018bert}). This has led to the development of several compressors using neural networks as predictors \cite{mahoney1, schmidhuber1996sequential, CLSTM}, including the recently proposed LSTM-Compress \cite{lstm-compress}, NNCP \cite{nncp} and DecMac \cite{decmac}. Most of the previous works, however, have been tailored for compression of certain data types (e.g., text \cite{decmac}\cite{cox2016syntactically} or images \cite{bitswap,townsend19}), where the prediction model is trained in a supervised framework on separate training data or the model architecture is tuned for the specific data type. This approach is therefore applicable only in the presence of existing training data and requires significant domain knowledge, and thus cannot be used for general-purpose compression. 

\subsection{Our Contributions}
In this work, we propose a general-purpose lossless compressor for sequential data, DZip, that relies on neural network based modeling. DZip treats the input file as a byte stream and does not require any additional training datasets. Hence, DZip is a standalone compressor capable of compressing any dataset (regardless of the alphabet size), unlike most existing neural network based compressors. We use a novel hybrid training approach which is ideally suited for such a setting and combines elements of adaptive and semi-adaptive modeling. 

We evaluate DZip on datasets from several domains including text, genomics and scientific floating point datasets, and show that it achieves on average 29\% improvement over Gzip \cite{gzip}, 33\% improvement over LSTM-Compress \cite{lstm-compress} (LSTM based compressor), and 12\% improvement over 7zip, reducing the gap between general-purpose and specialized compressors. DZip also outperforms the highly efficient lossless compressor BSC \cite{bsc} on most datasets with 8\% improvement on average, showing the advantages of improved modeling. In comparison to state-of-the-art NN-based compressors CMIX \cite{cmix} and NNCP, we demonstrate that DZip can achieve identical performance on most datasets of sufficiently large size while being 3-4 times faster than CMIX and 4 times faster than NNCP in encoding speed. Our results also indicate that for some datasets, the performance of DZip is close to that of specialized compressors, which are highly optimized for the specific datatypes.


In addition, we perform evaluations on certain synthetic datasets of known entropy that highlight the ability of the proposed compressor to learn long-term patterns better than the other general-purpose compressors. DZip serves as an example to showcase the potential of neural networks to boost compression performance. In contrast to traditional compressors, the current implementation of DZip suffers however from slower encoding and decoding speeds due to the complex neural network based modeling. However, compared to other neural network based compressors such as NNCP \cite{nncp} and CMIX \cite{cmix} (which uses thousands of context models followed by an NN based mixer), we show that DZip is several times faster. DZip is available as an open source tool at \url{https://github.com/mohit1997/Dzip-torch}, providing a framework to experiment with several neural network models and training methodologies.

\subsection{Related Works}
Ever since Shannon introduced information theory \cite{shannon1948} and showed that the \textit{entropy rate} is the fundamental limit on the compression rate for any stationary process, several attempts have been made to achieve this optimum. Several classes of general-purpose lossless compressors have been developed since then, including dictionary-based compressors (e.g., Gzip \cite{gzip}, 7zip/LZMA (\url{https://www.7-zip.org/})) and sorting transform based compressors (e.g., Bzip2 \cite{bzip2}, BSC \cite{bsc}). In addition, several specialized compressors have been developed, often using a statistical approach combining prediction models with arithmetic coding. For example, ZPAQ is a publicly available general purpose compressor which is specialized for text data. It uses several context models, context mixing, LZ77 coding, secondary symbol estimation and various other methods to improve compression performance while maintaining relatively fast compression and decompression speeds. ZPAQ handles text and non-text inputs differently, resulting in better performance on the former datasets.

Inspired by the performance of neural networks (NNs) in modeling and prediction tasks, several lossless compressors based on NNs have been proposed. \cite{schmidhuber1996sequential} proposed the application of a character-level recurrent neural network (RNN) model and showed competitive compression performance as compared to the existing compressors on text data. However, as vanilla RNNs were used, the performance was relatively subpar for complex sources with longer memory. More recently, LSTM-Compress \cite{lstm-compress} was proposed, which uses an LSTM (Long Short Term Memory Cells) model to adaptively learn the source distribution while encoding with an arithmetic coder. CMIX \cite{cmix}, the current state-of-the-art NN-based general-purpose lossless compressor, uses several thousand context models, most of which are based on PAQ8 (\url{http://mattmahoney.net/dc/}). This is further followed by an LSTM byte level mixer (to combine predictions) and a bit level NN based context mixer, which are trained through backpropagation adaptively while encoding the input data. NNCP \cite{nncp} is another RNN based compressor which adaptively compresses the input sequence while simultaneously updating the weights of the RNN. NNCP uses seven stacked LSTM layers which incorporate feature normalisation layers, further adding to the overall runtime. Moreover, the compressor only supports CPU based training and inference, resulting in extremely slow encoding speed.
There has also been work on designing specialized text compressors that exploit the generalization ability of NNs, using similar datasets for training the model to be used for compression (e.g., DecMac \cite{decmac} or \cite{cox2016syntactically} for text and BitSwap \cite{bitswap} or PixelVAE \cite{townsend19} for images). Most of these compressors are heavily specialized for a specific data type and require a model pretrained on similar data, thus limiting their applicability as general-purpose compression tools for arbitrary data types. 

In parallel to the work on compression, there has been significant progress in language modeling (e.g., BERT \cite{devlin2018bert}, GPT-2 \cite{radford2019language_gpt2}) and generative prediction models for images (e.g., PixelCNN++ \cite{salimans2017pixelcnn++}). In principle, these can be used for compression leading to significant improvements over the state-of-the-art, e.g., bringing the text compression rate below 1 bit per character. However, in practice, the model itself is typically quite large and needs vast amounts of data for training, which limits their direct applicability to general-purpose compression.

\section{Background}
Consider a data stream $S^N = \{S_1,S_2,\ldots, S_N\}$ over an alphabet $\mathcal{S}$ which we want to compress losslessly. We consider the statistical coding approach consisting of a prediction model followed by an arithmetic coder. For a sequence $S^N$, the aim of the model is to estimate the conditional probability distribution of the $r^{\textrm{th}}$ symbol $S_r$ based on the previously observed $K$ symbols, denoted as $P(S_r | S_{r-1},\ldots,S_{r-K})$, where $K$ is a hyperparameter. An estimate of this probability and $S_r$ are then fed into the arithmetic encoding block which recursively updates its state. This state serves for the compressed representation at the end of this process. The compressed size using this approach is equivalent to the cross entropy ($CE$) loss shown in Eq.\ \ref{eq:cat_cross}, where $|\mathcal{S}|$ is the alphabet size, $\underline{y}_r$, $\hat{\underline{y}}_r$ (vectors of size $|\mathcal{S}|$) are the one-hot encoded ground truth and the predicted probabilities, respectively, and $N$ is the sequence length. 
\begin{equation}\label{eq:cat_cross}
\mathcal{L} = \sum_{r=1}^N CE(\underline{y}_r, \underline{\hat{y}}_r) = \sum_{r=1}^N\sum_{k=1}^{|\mathcal{S}|} y_{rk}\log_2 \frac{1}{\hat{y}_{rk}}
\end{equation}
The model that estimates the probability $P(S_r | S_{r-1},\ldots,S_{r-K})$, where $r \in \{K+1, \ldots, N\}$, should be trained so as to minimize the cross entropy loss on the data to be compressed. This training can be performed in several ways \cite{surveycompression} as discussed below:\\

\begin{figure*}[!htbp]
	\centering
	\includegraphics[width=1\textwidth]{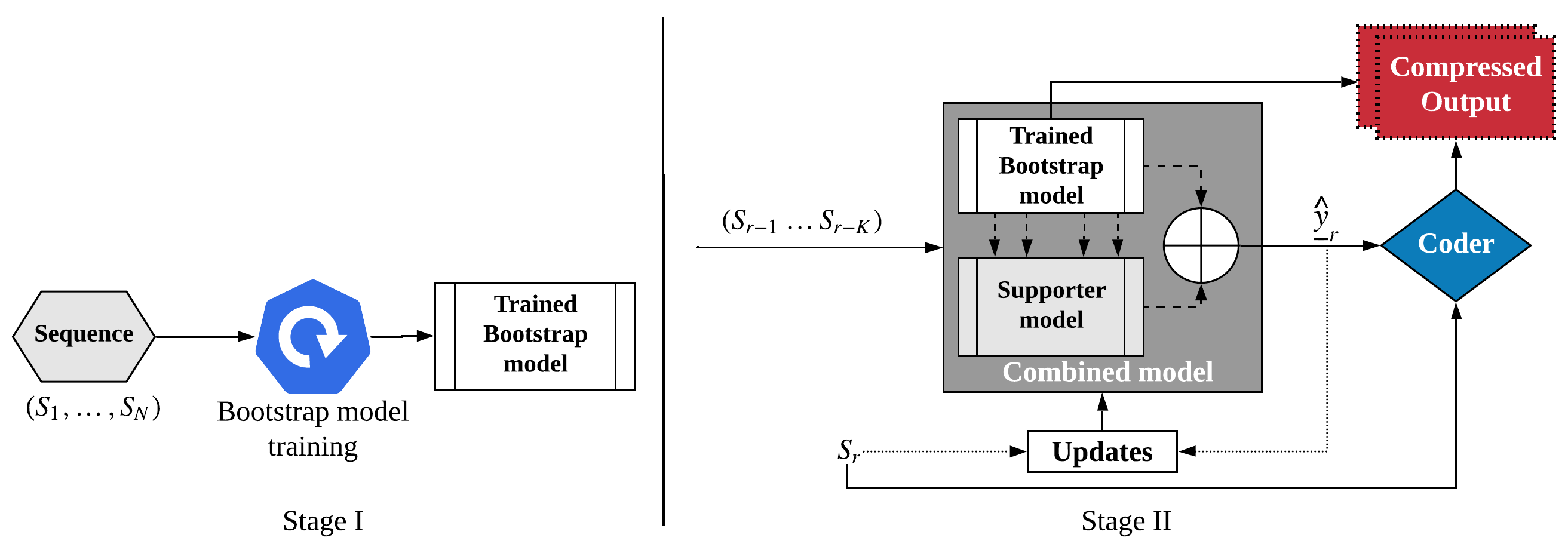}
	\caption{DZip compression overview: In Stage I, the boostrap model is trained by scanning the sequence multiple times. In Stage II, the bootstrap model is combined with the supporter model to predict the conditional probability of the current symbol given the past $K$ symbols ($K=64$ by default). The current symbol and the predicted probabilities are then fed into the arithmetic coder. The combined model is updated as the sequence is compressed. The final compressed output consists of the parameters of the trained bootstrap model and the output of the arithmetic coder.
	}
	\label{fig:algo}
\end{figure*}

\noindent
\textbf{Static}: Here the model is first trained on some external training data and it is made available to both the compressor and the decompressor. The performance in this case is highly dependent on the generalization abilities of the model. This approach is restricted to cases where similar training data is available and is not directly applicable to general-purpose compression tasks. \\

\noindent
\textbf{Adaptive}: Here both the compressor and the decompressor are initialized with the same random model which is updated adaptively based on the sequence seen up to some point. This approach does not require the availability of training data and works quite well for small models. For large models, however, this approach may suffer due to the difficulties in training the model in a single pass and adapting quickly to changing statistics (e.g., for non-stationary data).\\

\noindent
\textbf{Semi-adaptive}: Here the model is trained based only on the input sequence and the training procedure can involve multiple passes through the input data. The trained model parameters are included as part of the compressed file, along with the arithmetic coding output. This additional cost is expected to be compensated by the fact that the sequence-specific training will lead to a better predictive model and result in a smaller arithmetic coding output. Note that there is a trade-off involved between having an accurate model and the bits required to store that model's parameters, as described by the minimum description length (MDL) principle \cite{rissanen1978modeling}. Essentially, a larger model can lead to better compression, but the gains might be offset by the size of the the model itself, particularly for smaller datasets.  


In the next section we describe the proposed compressor DZip, which combines elements of semi-adaptive and adaptive approaches to achieve better prediction using NN-based models, while storing only a small model as part of the compressed file.

\section{Methods}

The proposed compressor DZip utilizes a hybrid training scheme that combines semi-adaptive and adaptive training approaches by means of two models, a \textit{bootstrap model} and a \textit{supporter model}, as shown in Figure \ref{fig:algo}. The bootstrap model is a parameter efficient RNN-based model that is trained in a \emph{semi-adaptive} fashion by performing multiple passes on the sequence to be compressed (prior to compression). Its parameters are saved and form part of the compressed output. The size of the bootstrap model is kept relatively small due to the trade-off associated with semi-adaptive modeling discussed above. 

To achieve further improvements in compression, we use the supporter model, which is a larger NN initialized with predefined pseudorandom parameters at the initiation of encoding and decoding. The outputs of the bootstrap and supporter models are combined to generate the final predictions used for compression. The parameters of the combined model are updated in an \emph{adaptive} manner during encoding (symmetrically during decoding). Due to the use of adaptive training, the weights of the supporter model do not need to be stored as part of the compressed file.

The parameter efficient bootstrap model provides a good initialization for the combined predictor, avoiding the issues with the adaptive training of large models. With this initialization, the larger combined model provides a powerful adaptively-trained predictor for large datasets, without incurring the cost associated with storing the parameters of the supporter model. The number of previous symbols used for prediction is set by default to $K=64$. As there exists a trade-off between compression performance and encoding/decoding speed, DZip also supports the use of the boostrap only model for prediction, avoiding the cost of adaptively training the large supporter model. We next describe the model architecture and the training procedure in more detail.

\begin{figure*}[htbp]
	\centering
	\includegraphics[width=\textwidth]{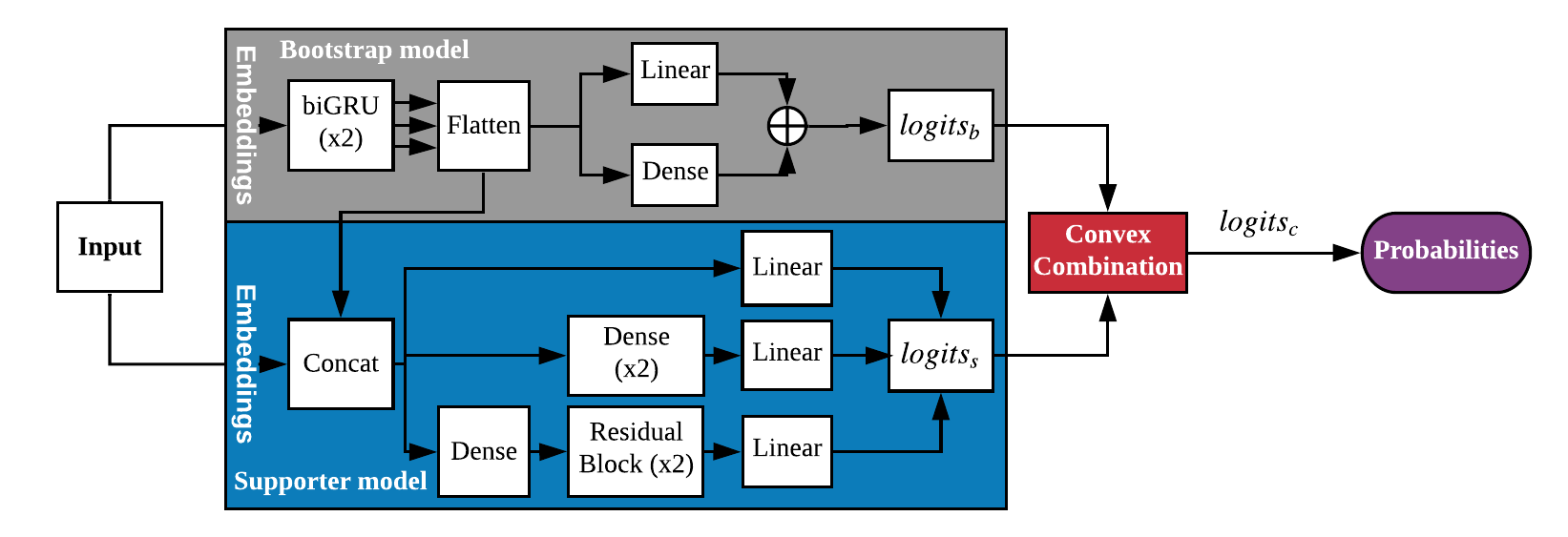}
	\caption{Combined model architecture consisting of bootstrap and supporter models. Dense layers correspond to fully connected layers with ReLU activation. Linear layers are also fully connected layers but do not incorporate a non linear transformation. Concat block denotes concatenation of all the input vectors.}
	\label{fig:aNN}
\end{figure*}

\subsection{Model architecture}

\textbf{Bootstrap model:} The bootstrap model architecture is designed keeping in consideration the trade-off between model size and prediction capability, leading to the choice of an RNN based design with parameter-sharing across time steps. The model is as shown in the top half of Figure \ref{fig:aNN} and consists of an embedding and two biGRU layers (bidirectional gated recurrent units \cite{GRU}) followed by linear and dense (fully connected) layers. The output of every $m^{\textrm{th}}$ time step after the biGRU layers is stacked and flattened out into a vector ($m=16$ by default). Choosing only the $m^{\textrm{th}}$ output helps in reducing the number of parameters in the next layer while still allowing the network to learn long-term dependencies. The small bottleneck dense layer (with ReLU activation) helps further increase the depth of the architecture and its output is added to that of the linear layer to generate the unscaled probabilities (logits) denoted as $logits_b$. This dense layer is important for learning long-term relationships in the inputs and showed improved performance on synthetic datasets. 

The layer widths of the bootstrap model are automatically chosen depending on the vocabulary size of the input sequence, since a higher vocabulary size demands larger input and output sizes. As the vocabulary size varies, the embeddings' dimensionality varies form 8 to 16; hidden state for biGRU varies from 8 to 128; and the dense layer's width (prior to logits) varies from 16 to 256. Note that the maximum possible alphabet size is 256 since the data is parsed at a byte level. The above hyperparameters were chosen empirically based on experimental results.\\

\textbf{Supporter model:} The supporter model architecture is designed to adapt quickly and provide better probability estimates than the bootstrap model, without any constraints on the model size itself. 
The input to this model consists of the embeddings and the intermediate representations from the bootstrap model (see Figure \ref{fig:aNN}). The supporter model consists of three sub-NNs which act as independent predictors of varying complexity. The first sub-NN is linear and learns quickly, the second sub-NN has two dense layers and the third sub-NN uses residual blocks \cite{resnet} for learning more complex patterns \cite{devlin2018bert}. We employ ReLU activation by default in all dense layers and the residual blocks. Then, each of the output vectors from these sub-NNs are linearly downsized into a vector of dimensionality equal to the vocabulary size and added together. The result of this operation is interpreted as the logits for the supporter model, denoted as $logits_{s}$. Based on empirical evaluation, the widths for the dense and residual layers are automatically set to 1024 or 2048 depending on the vocabulary size.\\

\textbf{Combined model:} The combined model takes the logits from the bootstrap model ($logits_b$) and the output logits from the supporter model ($logits_s$) to generate the final logits ($logits_c$) through a convex sum as shown below,
\begin{equation*}
    logits_c = \lambda*logits_b + (1-\lambda)*logits_s \; \textnormal{ s.t. } \; \lambda \in [0,1],
\end{equation*}
where $\lambda$ is a learnable parameter. We restrict the parameter $\lambda$ to $[0,1]$ through sigmoid activation. This allows the combined model to weigh the logits from the two models appropriately. At the initialisation, the prediction of supporter model can be expected to be poor, hence the convex combination allows the model to give more weight to $logits_b$ as compared to $logits_s$. The final output $logits_c$ is scaled to probabilities through softmax activation which are then used with arithmetic coding. This idea is similar to context mixing \cite{Adaptivecontext} which is commonly used to mix the predictions of multiple models resulting in superior compression performance.

\input{dataset_table}

\subsection{Model Training}
 The first stage of DZip involves model selection based on the vocabulary size of the input data, which is detected by doing a preliminary pass over the data. DZip reads the input file byte-by-byte and automatically selects hyperparameters for the bootstrap and the supporter model. The second stage consists of training the bootstrap model by performing multiple passes through the sequence. The model is typically trained for 8 epochs with a batch size of 2048, gradient clipping and Adam optimizer (learning rate 0.005) with learning rate decay while minimizing categorical cross entropy loss (Eq. \ref{eq:cat_cross}). For smaller datasets, the epoch number can be increased so as to sufficiently train the bootstrap model. 
After training, this model is saved as part of the compressed file after being losslessly compressed with general-purpose compressor BSC \cite{bsc}. Once the bootstrap model is trained, DZip can be used in two modes which trade-off compression ratio with encoding/decoding speed.\\

\noindent
\textbf{Combined Model (Hybrid)}: In this case, the prediction is done using the combined model, where the trained bootstrap model serves as a prior. The parameters for the supporter model are randomly initialized. During encoding and decoding, we symmetrically optimize the supporter model, while the bootstrap model's parameters are kept fixed. Since encoding one symbol at a time is extremely slow, we divide the sequence into 64 equally sized parts, and use a common model to generate predictions for each part in a single batch. This choice was made based on experimental results showing improved compression. After encoding a symbol $\underline{y}$ (represented as a one-hot vector), the parameters for the combined model are optimized by minimizing the following loss function:
\begin{equation*}
\mathcal{L}_{com} = CE(\underline{y}, f_s(logits_c)) + CE(\underline{y}, f_s(logits_s)),
\end{equation*}
where $f_s$ denotes the softmax activation, and $CE$ is the cross entropy loss defined earlier. The second term in this loss function forces the supporter model to learn even if the $logits_c$ are assigning more weight to the $logits_b$.
The weight updates are performed after encoding/decoding every 20 symbols (per part) while keeping the learning rate low (0.0005) to avoid divergence. To update this model, we use the Adam optimizer \cite{adam} with $\beta_1=0$ and $\beta_2=0.999$ to quickly adapt to the non-stationary sequence statistics. The hyperparameters such as the context length and the batch size used during adaptive encoding affect both the runtime and the compression performance of DZip. Details on the selection and the impact of these hyperparameters are provided in the supplementary data.
\\

\noindent
\textbf{Bootstrap Only Mode}: The sequence is divided into 1024 parts and the first $K$ symbols of each part are encoded using uniform probabilities. Note that the batch size used in this mode is 1024 as opposed to 64 for the combined modeling approach. Then we encode the $(K+1)^\textnormal{th}$ symbol in each part using the probability estimates obtained from the bootstrap model, where the prediction for each part is done in a single batch. This procedure is repeated until all parts are successfully encoded. The length of the sequence is stored as part of the encoded file. Decoding is performed in a symmetric fashion using the stored bootstrap model.

DZip uses the combined (hybrid) mode by default, as it offers better compression results, although at the cost of increased encoding/decoding times.

\subsection{Reproducibility}
The arithmetic coding procedure works symmetrically at the encoder and the decoder, and requires identical probability models at both ends for successful decoding of the data. We use the PyTorch guidelines on reproducibility \cite{pytorch17} to ensure identical training and inference during encoding and decoding. Since DZip utilizes a GPU to reduce runtime, its current implementation requires decoding to be performed on the same hardware that was used for encoding (a limitation of the PyTorch library).  DZip can be fairly easily adapted to the appropriate deep learning framework once reproducibility across GPUs becomes available. It is also possible to perform CPU based encoding/decoding which is reproducible across platforms, but is significantly slower. Note that the bootstrap model training can still be done on a GPU without any such concerns, since the trained model is included in the compressed file.

\input{results_real_data.tex}

\section{Experiments}
\label{sec:experiments}
We benchmark the performance of our neural network based compressor DZip on real and synthetic datasets, and compare it with state-of-the-art general-purpose compressors Gzip, BSC \cite{bsc}, 7zip \cite{7zip}, and ZPAQ \cite{zpaq}, as well as with RNN based compressors LSTM-Compress \cite{lstm-compress}, NNCP \cite{nncp} and CMIX \cite{cmix}. We also provide a comparison with specialized compressors for the real datasets when available. Certain neural network based compressors such as DecMac \cite{decmac} were not considered for comparison as they require a pretrained model or additional training data. Unless otherwise stated, all results corresponding to DZip are obtained using the combined model (default setting). DZip results are reported on a 16 GB NVIDIA Tesla P100 GPU and NNCP results are reported on 8 Intel(R) Xeon(R) CPU E5-2698 v4 @ 2.20GHz cores. 7zip, LSTM-Compress, BSC, ZPAQ, Gzip and CMIX results are reported on Intel Xeon Gold 6146 CPUs.

\subsection{Datasets}
\label{subsec: Datasets}

We consider a wide variety of real datasets with different alphabet sizes and sequence lengths, including genomic data (\textit{h. chr1, h. chr20, c.e. genome, np-bases, np-quality, ill-quality}), text (\textit{webster}\cite{silesia}, \textit{text8, enwiki9}), executable files {(\textit{mozilla}\cite{silesia})}, double precision floating point data (\textit{num-control, obs-spitzer, msg-bt}), and audio data (\textit{audio} \cite{ESC50}). To further understand the capabilities of DZip, we also test it on synthetic datasets \cite{knuth2014art} with known entropy rate and increasing complexity. See Table \ref{table:data} for a detailed description. Links to the considered datasets are provided on the GitHub repository \url{https://github.com/mohit1997/Dzip-torch}.


\subsection{Results on real data}
\label{subsec: comp real data}

 We first analyze the performance of DZip on the real datasets (see Table \ref{table:results_real}). On each dataset, we include results for specialized compressors (when available) as their performance serves as a baseline for achievable compression. In particular, we use CMIX \cite{cmix} for \textit{webster, mozilla, text8}, and \textit{enwiki9}, GeCo \cite{geco} for \textit{h.\ chr20, h.\ chr1, c.e.\ genome}, and \textit{np-bases}\footnote{GeCo is a specialized compressor for genomic sequences, and is not optimized for nanopore genomic read bases.}, DualCtx \cite{y2019compression} for \emph{np-quality}\footnote{By default, DualCtx uses read bases as an additional context for quality value compression. However, we do not use the read base context to allow fair comparison with other tools.}, QVZ \cite{malysa2015qvz} for \textit{ill-quality}\footnote{QVZ is optimized for lossy compression, but also provides a mode for lossless compression.}, and FPC \cite{FPC} for \emph{msg-bt, num-control}, and \emph{obs-spitzer} datasets.  Since ZPAQ is also specialized for text data, we discuss its comparison with DZip on \textit{webster, text8} and \textit{enwiki9}.  We do not experiment with any specialized compressors for \emph{audio} dataset since we generate the binary audio file by concatenating multiple audio files together.

\input{results_synthetic.tex}

When compared against Gzip, BSC, 7zip, and LSTM-Compress, DZip offers the best compression performance for all datasets except for the \emph{webster} and \emph{mozilla}, in which BSC, LSTM-Compress or 7zip perform better than DZip. On \emph{np-quality} and \emph{text8} datasets, the bpc achieved by DZip is comparable to BSC which outperforms Gzip, LSTM-Compress and 7zip on most of the datasets. DZip, on average, offers about 29\% improvement over Gzip, 12\% improvement over 7zip, 9.4\% improvement over BSC and 33\% improvement over LSTM-Compress. While ZPAQ, which is specialized for text datasets, consistently outperforms DZip on \emph{webster, mozilla, text8} and \emph{enwiki9}, this improvement becomes smaller as the length of the sequence increases. For example, for \emph{mozilla} dataset (length 51.2M), the difference in bpc is 0.27 and for \emph{enwiki9} (length 500M) the difference is reduced to 0.04 bpc. On other datasets, DZip always outperforms ZPAQ (except for \emph{np-quality} and \emph{c.e. genome} where the difference is small). We also observe that the performance of LSTM-Compress varies significantly across datasets, in some cases performing worse than the bpc (bits per character) achievable using an independent uniform distribution. This might be attributed to the issues with training LSTMs (exploding and vanishing gradients) and also the hyperparameters such as the learning rate which cannot be externally controlled in the case of LSTM-Compress.

When compared with NNCP, we observe that NNCP achieves better performance on \textit{webster, mozilla, text8, enwiki9} and \textit{obs-spitzer} with around 20\% improvement over DZip. NNCP performs better than DZip on the text and executable datasets, while the performance on the remaining datasets is near-identical. On a few datasets such as \textit{msg-bt} and \textit{audio}, DZip provides slight improvement over NNCP. Finally, CMIX was observed to perform consistently better than all other compressors on all datasets except \textit{msg-bt} and \textit{audio} where DZip performs slightly better (0.03 and 0.04 improvement in bpc, respectively). Since CMIX is specialized for text and executable files, the bpc is significantly better for \textit{webster, mozilla, text8} and \textit{enwiki9} datasets. On other datasets, CMIX on average provided approximately 3-4\% improvement over DZip. Note that CMIX is expected to outperform DZip, as CMIX uses several thousand context models which are specialized for specific data types. In contrast, DZip uses a single model and does not incorporate any data type specific knowledge. Nevertheless, the results of DZip demonstrate that deep learning based lossless compression techniques can perform comparably to CMIX (while being faster, as shown in subsection \ref{subsec:runtime}) on datatypes such as audio, floating point and genomic data, for which CMIX has not been specialized.

It is important to note that the performance of DZip is sensitive to the alphabet size and the sequence length because of the overhead associated with storing the bootstrap model parameters. Specifically, for small length and large alphabet sequences, the bootstrap model size contributes a significant percentage to the overall compressed size, hurting the overall compression ratio. This is reflected for example in the \emph{webster} and \emph{mozilla} datasets, where the model occupies 31\% and 25\% of the compressed file size, respectively, resulting in worse compression performance than LSTM-Compress, 7zip and BSC. However, as the sequence length increases, DZip outperforms BSC, LSTM-Compress, and 7zip on all datasets as the model size contribution gets amortized. For example, on \emph{enwiki9} dataset, which is 5 times larger than \emph{text8}, DZip obtains 10\% improvement over BSC and LSTM-Compress, and gives comparable performance to ZPAQ which is specialized for this data type. With the increase in sequence length, DZip is also able to achieve a performance close to that of NNCP and CMIX, such as on \textit{np-bases, np-quality, msg-bt} and \textit{audio} datasets. However, this might not always be the case, as reflected in \textit{obs-spitzer} dataset which is of comparable length. 
Note that this analysis serves to compare DZip more fairly to other methods that do not need to store the model. As discussed, storing the model parameters incurs a cost in the overall achieved bpc, specially for small datasets, but on the other hand, it allows for much faster decoding (see subsection \ref{subsec:runtime}), which is an important practical consideration.

Focusing on the specialized compressors, we observe that they outperform general-purpose compressors in all cases except for the genomic files \emph{ill-quality, np-bases, msg-bt, num-control} and \emph{obs-spitzer}. Regarding the genomic files, this result may be explained because QVZ and GeCo are not optimized for lossless compression of quality data and nanopore bases, respectively. Similarly, FPC (used to compress \emph{msg-bt, num-control} and \emph{obs-spitzer}) is optimised for encoding/decoding speed rather than compression performance. When compared to DZip on the other genomic datasets, GeCo provides less than 2\% reduction on \emph{Human} chromosomes and 5\% reduction on the \textit{C. Elegans genome}. Similarly, for the \textit{webster, text8} and \emph{enwiki9} datasets, ZPAQ results in 22\%, 12\% and 5\% improvement over DZip, respectively, while CMIX provides  42\%, 24\% and 28\% improvement over DZip, respectively. Moreover, for \emph{mozilla} dataset, CMIX provides 35\% improvement as compared to DZip. These results are expected, since the specialized compressors typically involve handcrafted contexts and mechanisms which are highly optimized for the particular dataset statistics and are based on large training datasets. Nevertheless, DZip achieves a performance reasonably close to that of the specialized compressors. The gap is further reduced if the model size is not taken into account, even outperforming the specialized compressors in some cases. For example, in comparison to ZPAQ, the compression rate for DZip without the model size would be 10\% lower for \emph{webster}, similar for \emph{enwik9}, and only 3\% larger for \emph{text8}.

\subsection{Results on synthetic data}
We further evaluate DZip on synthetic datasets with simple structure (i.e., low Kolmogorov complexity) but long-term dependencies, which make them difficult to compress using traditional compressors. Specifically, we tested on two sequence classes with known entropy rates, \textit{XOR-k} (entropy rate 0) and \textit{HMM-k} (entropy rate 0.469), where \textit{k} represents the memory of the sequence (see Table \ref{table:data}). 

Table \ref{table:results_synthetic} shows the results for increasing values of \textit{k} for the two sequence classes. Note that these are binary sequences and hence it is possible to achieve 1 bpc just by packing the sequence in bits. We observe that DZip achieves the best compression performance in all cases, almost achieving the entropy rate of the corresponding sequences when $k < 70$, with slight overhead due to the bootstrap model size. Note that DZip uses 64 previous symbols for prediction, making it impossible to learn dependencies beyond this range. While this hyperparameter can be increased by the user, it would result in slower encoding/decoding speeds reducing the practicality of our method.
Also note that Gzip, BSC, ZPAQ and LSTM-Compress fail to achieve any meaningful compression, except for $k=20$ in which case BSC and ZPAQ are able to capture the dependency to some extent. For example, Gzip requires 1.2 bpc in all cases, and LSTM-Compress employs more than 3 bpc except for the \textit{HMM-70} dataset. 7zip on the other hand gives lower than one bpc for \textit{XOR-k} datasets, but its performance is still worse than DZip for $k < 70$. However, on \textit{HMM-k} datasets, 7zip is also unable to capture the dependency in the input sequence. Similarly, NNCP also fails to learn the relationships, resulting in poor compression performance on all synthetic data-sets for $k>20$. Note that RNNs generally face difficulties capturing these long term dependencies \cite{longtermRNN} and do not scale well to large context sizes. Lastly, CMIX gives similar trends in compression as the other compressors and only gives meaningful compression for sequences with $k=20$.

\input{results_comparison}

\input{runningtime_table}

\subsection{Tradeoff between ``bootstrap only'' and combined modeling approaches}
To understand the benefits of the combined modeling approach adopted by DZip, we conduct ablation experiments where we compare the two modes for DZip: (i) compression using only the trained bootstrap model and (ii) compression using the combined DZip hybrid model with adaptive training (default setting). Table \ref{table:results_comparison} shows the results for the real datasets and two synthetic datasets that serve as  representative datasets. On average, we observe that using the proposed combined model improves the compression by 0.055 bpc on the real datasets. No significant difference in performance is observed on the selected synthetic datasets between the two approaches. This improvement is obtained at the cost of higher encoding and decoding time, since the combined model is more complex and needs to be adaptively trained. In particular, as shown in Table \ref{tab:runtime}, the encoding and decoding speed is on average 1.5 and 4 times faster for the bootstrap only mode. Nevertheless, DZip in bootstrap only mode still outperforms Gzip, 7zip, BSC and LSTM-Compress on most of the selected datasets, while being more practical due to its reduced running time. For example, we observe 27\% improvement with respect to Gzip, 31\% improvement with respect to LSTM-Compress, 9\% improvement with respect to 7zip,  and 6\% improvement over BSC. Hence, depending on speed and performance requirements, one mode may be preferred over the other.

Finally, we observe that the improvement provided by the combined model varies across the different datasets. Specifically, the improvement is smaller on \textit{ill-quality, np-quality, num-control} and \textit{audio} datasets. The reason might be that the bootstrap model is already close to the compression limit of the sequence. This is the case for example on the synthetic datasets, where the performance of the bootstrap model is already close to the entropy rate and hence the improvement offered by the hybrid (combined) model is negligible.

\subsection{Computational requirements}
\label{subsec:runtime}
The first stage of DZip, which consists on training the bootstrap model with a batch size of 2048, employs 2-5 minutes/MB (depending on the alphabet size). We typically train the bootstrap model on every dataset for 8 epochs. In the bootstrap only mode, the encoding and decoding typically take 0.4 minutes per MB and 0.7 minutes per MB, respectively. With the combined modeling approach, since it requires weight updates after every 20 symbols, the encoding time rises to roughly 2.5 minutes per MB. The decoding speed is 15\% slower than encoding speed in this case due to the method used for creating batches in the current implementation. A detailed run time analysis is provided in Table \ref{tab:runtime} for various vocabulary sizes evaluated in this work.



DZip outperforms other NN-based compressors in terms of computational performance because it relies on simpler models and it uses GPU along with various parallelisation schemes adopted during both training and encoding. LSTM-Compress takes on average 3 minutes/MB for encoding and 4 minutes/MB for decoding. Therefore, LSTM-Compress is 2 times faster in encoding speed and 5 times slower in decoding speed than DZip in bootstrap only mode. Compared to DZip in combined mode, LSTM-Compress is 3 times faster in encoding speed and 1.3 times slower in decoding speed. Note that while LSTM-Compress gives somewhat comparable encoding/decoding speeds, DZip consistently outperforms LSTM-Compress on the majority of datasets in terms of compression ratio. Compared to NNCP, DZip in bootstrap only mode compresses 5-6 times faster and decompresses 60 times faster. In combined mode, DZip compresses 3-4 times faster and decompresses 15 times faster than NNCP. CMIX uses specific preprocessing transformations based on the data type and has variable encoding/decoding speed. On average, DZip in combined mode is more than 4 times/10 times faster in encoding/decoding speed than CMIX. CMIX averages around 20-32 minutes/MB even with its highly optimized implementation for both compression and decompression. In bootstrap only mode, DZip is 5 times faster for compression and 25 times faster for decompression than CMIX.

In comparison, Gzip, 7zip and BSC take on average 4.9 seconds/MB, 0.13 seconds/MB, and 0.07 seconds/MB for compression, respectively, and 0.005 seconds/MB, 0.04 seconds/MB and 0.02 seconds/MB for decompression, respectively. ZPAQ's compression speed is variable for different datasets and different modes. In the mode used for the experiments (mode five), ZPAQ requires approximately 1-2 seconds per MB for encoding and around 5-10\% more time for decompression. The difference in compression speeds is expected since training and inference for NNs are expensive, but they can provide better compression rates due to superior modeling capabilities. 

\section{Conclusion}
In this work, we introduce a general-purpose neural network prediction based framework for lossless compression of sequential data. The proposed compressor DZip uses a novel NN-based hybrid modeling approach that combines semi-adaptive and adaptive modeling. We show that DZip achieves improvements over Gzip, 7zip, BSC, ZPAQ and LSTM-Compress for a variety of real datasets and near optimal compression for synthetic datasets. DZip also compares favorably with the other NN-based compressors, achieving similar compression while being substantially faster. Although the practicality of DZip is currently limited due to the required encoding/decoding time, we believe the proposed framework and experiments can shed light into the potential of neural networks for compression, as well as serve to better understand the neural network models themselves.

Future work includes reducing the computational requirements to build a more practical tool, improved compression of the trained bootstrap model parameters, and support for incorporating domain specific knowledge when available.

\bibliographystyle{IEEEtran}
\bibliography{IEEEabrv,refs}
%



%

\begin{IEEEbiography}[{\includegraphics[width=1in,height=1.25in,clip,keepaspectratio]{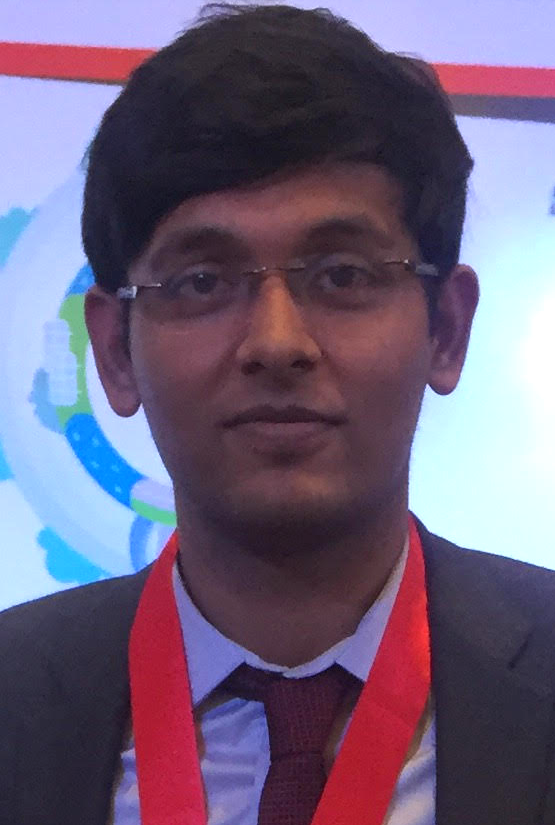}}]{Mohit Goyal} received the B. Tech. degree in electrical engineering from IIT Delhi, Delhi India, in 2019. He is currently pursuing the M.S. degree in electrical and computer engineering from University of Illinois at Urbana Champaign. His research interests include data compression, neural networks, machine learning and bioinformatics.
\end{IEEEbiography}

\begin{IEEEbiography}[{\includegraphics[width=1in,height=1.25in,clip,keepaspectratio]{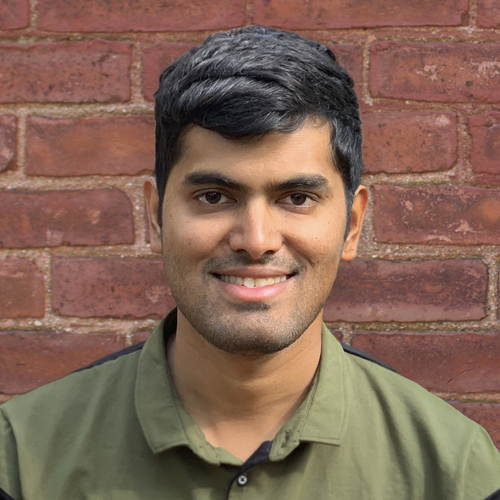}}]{Kedar Tatwawadi} received his Ph.D. from Stanford University, where he specialized problems using machine learning for compression and error correction coding. He holds a B.Tech in Electrical Engineering Indian Institute of Technology, Bombay, and a M.S. from Stanford University. Kedar is the recipient of the Numerical Technologies Founders Prize at Stanford, and the Qualcomm Innovation Fellowship.
\end{IEEEbiography}


\begin{IEEEbiography}[{\includegraphics[width=1in,height=1.25in,clip,keepaspectratio]{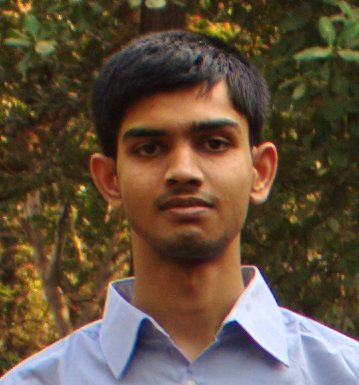}}]{Shubham Chandak} received the B.Tech. degree in electrical engineering from IIT Bombay, Mumbai, India, in 2016, and the M.S. degree in electrical engineering from Stanford, CA, USA in 2018, where he is currently pursuing the Ph.D. degree. His current research interests include data compression, DNA storage, information theory, and machine learning.

\end{IEEEbiography}

\begin{IEEEbiography}[{\includegraphics[width=1in,height=1.25in,clip,keepaspectratio]{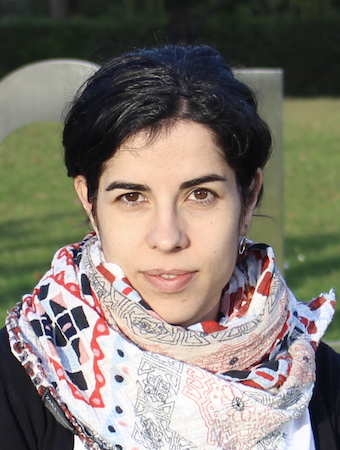}}]{Idoia Ochoa}
is an Assistant Professor of the Electrical and Computer Engineering Department
at the University of Illinois at Urbana-Champaign (UIUC), since January 2017, and a visiting professor of the School of Engineering TECNUN, at the University of Navarra, in Spain.
In 2016 she obtained a Ph.D.~from the Electrical Engineering Department at Stanford University.
She received her M.Sc.~from the same department in 2012.
Prior to Stanford, she graduated with B.Sc.~and M.Sc.~degrees in Telecommunication Engineering
(Electrical Engineering) from the University of Navarra, Spain, in 2009.
She is recipient of the MIT Innovators under 35 award and the Stanford Graduate Fellowship.
Her research interests include data compression, computational biology, bioinformatics, and machine learning.
\end{IEEEbiography}


\vfill


\end{document}

%% file: dataset_table.tex
\begin{table*}[!htbp]
	\centering
	\caption{Real and synthetic datasets used for evaluation. {$|\mathcal{S}|$} denotes the alphabet size. bpc stands for bits per character. 
	}
	\label{table:data}
	\vskip 0.15in
	\begin{small}
		\begin{tabular}{llll}
			\hline
			\textbf{Name}     & \textbf{Length} & {$|\mathcal{S}|$} & \textbf{Description} \\ 
			\hline
			\textbf{Real Datasets} & & &\\
			\hline
			
			a) Text & & &\\
			\hline
			\emph{webster}& 41.1M & 98 & HTML data of the 1913 Webster Dictionary, from the Silesia corpus \\
			\emph{text8} & 100M & 27 &  First 100M of English text (only) extracted from \emph{enwiki9}\\
			\emph{enwiki9}&  500M & 206 & First 500M of the English Wikipedia dump on 2006\\
			\hline
			
			b) Executables  & & &\\
			\hline
			\emph{mozilla}& 51.2M & 256 & Tarred executables of Mozilla 1.0, from the Silesia corpus\\
			\emph{h.\ chr20} & 64.4M & 5 &  Chromosome 20 of \emph{H. sapiens} GRCh38 reference sequence 
			\\
			\hline
			c) Genomic Data & & &\\
			\hline
			\emph{h.\ chr1} & 100M & 5 & First 100M bases of chromosome 1 of \emph{H. Sapiens} GRCh38 sequence
			\\
			\emph{c.e.\ genome}  & 100M & 4 & \textit{C.\ elegans} whole genome sequence\\
			\emph{ill-quality}& 100M & 4 & 100MB of quality scores for PhiX virus reads sequenced with Illumina \\
			
			\emph{np-bases}&  300M & 5 & Nanopore sequenced reads (only bases) of a human sample (first 300M symbols) \\
			\emph{np-quality}& 300M & 91 & Quality scores for nanopore sequenced human sample (first 300M symbols)\\
			\hline
			d) Floating Point & &\\
			\hline
			\emph{num-control}&  159.5M & 256 & Control vector output between two minimization steps in weather-satellite data assimilation\\
			\emph{obs-spitzer}&  198.2M & 256 & Data from the Spitzer Space Telescope showing a slight darkening\\
			\emph{msg-bt}&  266.4M & 256 & NPB computational fluid dynamics pseudo-application bt\\
			\hline
			
			e) Audio (wav) & & &\\
			\hline
			\emph{audio}&  264.6M & 256 & First 600 files (combined) in ESC Dataset for environmental sound classification \\
			\hline
			\textbf{Synthetic Datasets} & & &\\
			\hline
			\emph{XOR-k}  & 10M & 2 & Pseudorandom sequence $S_{n+1} = S_n + S_{n-k}\ (\textnormal{mod}\ 2)$. \\
			&&& Entropy rate $0$ bpc.\\
			\emph{HMM-k} & 10M & 2 & Hidden Markov sequence $S_n = X_n + Z_n\ (\textnormal{mod}\ 2)$, with $Z_n \sim Bern(0.1)$,\\
			& & & $X_{n+1} = X_n + X_{n-k}\ (\textnormal{mod}\ 2)$. Entropy rate $0.46899$ bpc. \\
			\hline
	\end{tabular}
	\end{small}
\end{table*}

%% file: results_real_data.tex
\begin{table*}[!htbp]
\centering
\caption{Bits per character (bpc) achieved by the tested compressors on the real datasets. Best results are boldfaced. $\log_2|\mathcal{S}|$ represents the bpc achieved assuming an independent uniform distribution over the alphabet of size $|\mathcal{S}|$. For DZip, we specify the total bpc and the size of the model (in \% space occupied). Spec. Comp. stands for specialized compressor. 
}
\label{table:results_real}
\vskip 0.15in
\begin{small}
\begin{sc}
\begin{tabular}{ccccccccc|cc|c}
\toprule
\multirow{2}{*}{\textbf{File}} & \multirow{2}{*}{\textbf{Len/$\mathbf{log_2|\mathcal{S}|}$}} & \multirow{2}{*}{\textbf{Gzip}} & \multirow{2}{*}{\textbf{BSC}} & \multirow{2}{*}{\textbf{7zip}} & \multirow{2}{*}{\textbf{ZPAQ}} & \multicolumn{1}{c}{\textbf{LSTM}}  & \multirow{2}{*}{\textbf{NNCP}} & \multicolumn{1}{l|}{\multirow{2}{*}{\textbf{CMIX}}} & \multicolumn{2}{c|}{\textbf{DZip}}                                                           & \multirow{1}{*}{\textbf{Spec.}} \\ \cline{10-11}
                               &                                     &                                                               & & & & \multicolumn{1}{c}{\textbf{Compress}} & & \multicolumn{1}{l|}{}                              & \multicolumn{1}{c|}{\textbf{bpc}} & \multicolumn{1}{l|}{\textbf{Model}} &  \multirow{1}{*}{\textbf{Comp.}}                               \\ \midrule
\textit{webster}               & 41.1M/6.61                            & 2.32                                                                 & 1.29             & 1.70  & 1.09   & 1.23 &   0.98  &   \textbf{0.83}                         & 1.44                                & 31.33\%                                  & 0.83                            \\
\textit{mozilla}               & 51.2M/8.00                           & 2.97                                                                & 2.52                & 2.11   & 1.88 & 2.05 &  1.63  &   \textbf{1.39}                         & 2.15                                & 25.37\%                                & 1.39                              \\
\textit{h.\ chr20}              & 64.4M/2.32                             & 2.05                                                                           & 1.73             & 1.77  &    1.68 & 7.82   & 1.66    &    \textbf{1.62}                     & 1.63                       & 0.92\%                                  & 1.62                            \\
\textit{h.\ chr1}               & 100M/2.32                            & 2.14                                                                           & 1.78         & 1.83 &   1.74  & 7.36  &   \textbf{1.67}  &       \textbf{1.67}                         & \textbf{1.67}                       & 0.58\%                                  & 1.65                            \\
\textit{c.e.\ genome}              & 100M/2.00                              & 2.15                                                                          & 1.87         & 1.89    &  1.80  & 7.51  &  1.80   &     \textbf{1.74}                          & 1.81                       & 0.53\%                                 & 1.72                            \\
\textit{ill-quality}             & 100M/2.00                              & 0.50                                                                           & 0.35          & 0.35      & 0.34 & 6.48 &   0.34  &    \textbf{0.33}                         & 0.34                       & 2.79\%                                   & 0.51                            \\
\textit{text8}                 & 100M/4.75                             & 2.64                                                                          & 1.68        & 1.93  &  1.52  & 1.76 &  1.48   &    \textbf{1.31}                     & 1.74                                & 9.38\%                                      & 1.31                            \\
\textit{np-bases}              & 300M/2.32                              & 2.16                                                                           & 1.86               & 1.93 &  1.79 & 7.34   &   \textbf{1.70}  &      1.73                       & 1.73                       & 0.19\%                                 & 1.75                            \\
\textit{np-quality}              & 300M/6.51                             & 5.95                                                                           & 5.69     & 5.71   & 5.53 & 5.51   &   5.50  &     \textbf{5.49}                                & 5.56                       & 1.13\%                                  & 5.35                             \\ 
\textit{enwiki9}               & 500M/7.69                            & 2.72                                                                           & 1.64    & 1.94        &     1.43  & 1.66   &   1.21  &    \textbf{1.05}                    & 1.47                       & 3.67\%                                     & 1.05                            \\

\textit{num-control}               & 159.5M/8.00                            & 7.57                                                                           & 7.66    & 7.41     &   6.96   & 6.82   &   6.72  &    \textbf{6.63}                    & 6.83                 & 2.67\%                                     & 7.12                            \\
\textit{obs-spitzer}               & 198.2M/8.00                            & 6.50                                                                           & 2.51   & 2.27      &     2.20   & 2.87         &   1.73  &    \textbf{1.58}                   & 2.18                 & 6.70\%                                     & 7.74                            \\
\textit{msg-bt}               & 266.4M/8.00                            & 7.08                                                                           & 6.96     & 5.76          &    6.29   & 6.22     &   5.36  &    5.24                    & \textbf{5.21}                       & 2.08\%                                     & 6.67                            \\
\textit{audio}               & 264.6M/8.00                            & 5.75                                                                           & 4.63   &   4.98           &    4.17  & 4.92   &   3.49  &    3.44                    & \textbf{3.40}                       & 3.29\%                                     & N/A                            \\\bottomrule
\end{tabular}
\end{sc}
\end{small}

\end{table*}

%% file: results_synthetic.tex
\begin{table*}[!htbp]
\centering
\caption{Bits per character for synthetic datasets. Best results are boldfaced. Note that 1 bpc can be achieved simply by using 1 bit per symbol for binary sequences.}
\label{table:results_synthetic}
\begin{small}
\begin{sc}
\begin{tabular}{ccccccccc}
\toprule
 \textbf{Compressor} & \textit{XOR-20} & \textit{XOR-30} & \textit{XOR-50} & \textit{XOR-70}   & \textit{HMM-20} & \textit{HMM-30} & \textit{HMM-50} & \textit{HMM-70} \\ 
 \midrule
Gzip   & 1.20          & 1.20          & 1.19          & 1.19            & 1.19          & 1.19          & 1.19          & 1.19          \\ 
BSC    & 0.10          & 1.01          & 1.01          & 1.01            & 0.69          & 1.01          & 1.01          & 1.01         \\ 
7zip  & 0.05          & 0.69          & 0.65            & \textbf{0.62}          & 0.92          & 1.03          & 1.02      & 1.03          \\ 
ZPAQ   &      0.08     & 0.95          & 1.00          & 1.00            &    0.99       & 0.99          & 0.99         & \textbf{0.99}         \\
LSTM-Compress   & 4.23          & 3.19          & 4.77          & 3.43            & 3.02          & 5.19          & 3.64          & 1.01         \\ 
NNCP   &      0.125     & 0.9          & 1.00          & 1.00            &    0.57       & 1.00          & 1.00         & 1.00         \\
CMIX   &      0.02     & 0.87          & 0.93          & 0.96 &    0.50       & 0.85          & 0.99         & \textbf{0.99} \\
DZip   & \textbf{0.01} & \textbf{0.01} & \textbf{0.01} & 1.01 & \textbf{0.48} & \textbf{0.48} & \textbf{0.48} & 1.01 \\
\bottomrule
\end{tabular}
\end{sc}
\end{small}
\end{table*}


%% file: results_comparison.tex

\begin{table}[!htbp]
\centering
\caption{Compression in bpc obtained by (i) only the bootstrap model and (ii) DZip (combined model). Improv. stands for the improvement of the combined model with respect to the bootstrap model (in bpc).}
\label{table:results_comparison}
\vskip 0.15in
\begin{small}
\begin{sc}

\begin{tabular}{cccccc}
\toprule
\multirow{2}{*}{\textbf{File}}     & \multirow{2}{*}{\textbf{Length}} & \textbf{Bootstrap}  & \multirow{2}{*}{\textbf{DZip}} & \multirow{1}{*}{\textbf{Improv.}} \\ 
     &  & \textbf{only} & & \textbf{(bpc)}\\ \midrule
\textit{webster}  & 41.1M & 1.474 &  1.443  & 0.031 \\ 
\textit{mozilla}  & 51.2M & 2.233 &  2.150  & 0.083 \\ 
\textit{h. chr20} & 64.4M & 1.672 &  1.634 & 0.038\\ 
\textit{h. chr1} & 100M & 1.720 & 1.673 & 0.047\\ 
\textit{c.e. genome} & 100M & 1.826 &  1.814 & 0.012\\ 
\textit{ill-quality} & 100M & 0.343 & 0.343 & 0.000 \\ 
\textit{text8} & 100M & 1.789 & 1.737 & 0.052\\ 
\textit{np-bases} & 300M & 1.755 & 1.725 &0.03\\ 
\textit{np-quality} & 300M & 5.588 & 5.562 & 0.026\\ 
\textit{enwiki9} & 500M & 1.596 & 1.470 &0.126\\
\textit{num-control} & 159.5M & 6.838 & 6.834 &0.004\\
\textit{obs-spitzer} & 198.2M & 2.445 & 2.181 &0.264\\
\textit{msg-bt} & 266.4M & 5.259 & 5.214 &0.045\\
\textit{audio} & 264.6M & 3.405 & 3.389 &0.016\\\midrule
\textit{XOR-30} & 10M & 0.011 & 0.011 & 0.0\\ 
\textit{HMM-30} & 10M & 0.482 & 0.482 & 0.0\\ 
\bottomrule
\end{tabular}
\end{sc}
\end{small}
\end{table}

%% file: runningtime_table.tex
\begin{table*}[!htbp]
\centering
\caption{Running time of DZip in Minutes per MB for various alphabet sizes. The decoding speed with the combined model is 15\% slower than encoding. For bootstrap model the encoding and decoding speeds are similar. The bootstrap training times shown are for 8 epochs.}
\label{tab:runtime}
\vskip 0.15in
\begin{small}
\begin{sc}
\begin{tabular}{ccccccccc}
\toprule
\textbf{Step} & \textbf{$|\mathcal{S}|$=2} & \textbf{$|\mathcal{S}|$=4} & \textbf{$|\mathcal{S}|$=5} & \textbf{$|\mathcal{S}|$=27} & \textbf{$|\mathcal{S}|$=91} & \textbf{$|\mathcal{S}|$=98} & \textbf{$|\mathcal{S}|$=206} & \textbf{$|\mathcal{S}|$=256} \\ \midrule
\textit{Bootstrap Training}      & 1.5        & 2.0        & 2.0        & 5.2         & 5.2         & 5.2         & 5.5          & 5.5          \\ 
\textit{Bootstrap Encoding}     & 0.3        & 0.3        & 0.3        & 0.4         & 0.4         & 0.4         & 0.6          & 0.7          \\ 
\textit{Bootstrap Decoding}     & 0.5        & 0.5        & 0.5        & 0.7         & 0.7         & 0.7         & 0.9          & 1.1          \\ 
\textit{Combined Encoding}      & 1.9        & 2.2        & 2.2        & 2.4         & 2.4         & 2.4         & 2.7          & 2.9          \\
\textit{Combined Decoding}      & 2.0        & 2.5        & 2.5        & 2.7         & 2.7         & 2.7         & 3.0          & 3.3          \\
\bottomrule
\end{tabular}
\end{sc}
\end{small}
\end{table*}